\documentclass[letterpaper, 10 pt, conference]{ieeeconf}  
\usepackage{cite}
\usepackage{graphicx}

\usepackage{cuted}
\usepackage{caption}

\usepackage{amsfonts}   
\usepackage{amsmath}

\usepackage{booktabs}

\usepackage{hyperref}
\hypersetup{
    colorlinks=true,      
    linkcolor=blue,       
    filecolor=magenta,     
    urlcolor=cyan,         
}

\IEEEoverridecommandlockouts                              

\title{\LARGE \bf
ABPolicy: Asynchronous B‑Spline Flow Policy for Real‑Time and Smooth Robotic Manipulation
}

\author{Fan Yang$^{1}$, Peiguang Jing$^{1}$, Kaihua Qu$^{1}$, Ningyuan Zhao$^{1}$, and Yuting Su$^{1,*}$
\thanks{$^{*}$Corresponding author.}
\thanks{$^{1}$The authors are with the School of Electrical and Information Engineering, Tianjin University
        { (e-mail: eeyf@tju.edu.cn; pgjing@tju.edu.cn; 1106153048@qq.com; nyzhao@tju.edu.cn; ytsu@tju.edu.cn).}}
\thanks{*This work was supported by the Guangxi Natural Science Foundation Key Project under Grant 2025JJD170006, and the National Natural Science Foundation of China under Grant 62361002.}
}

\begin{document}

\maketitle
\thispagestyle{empty}
\pagestyle{empty}


\begin{strip}
\centering
\includegraphics[width=\textwidth, trim=10 270 78 8, clip]{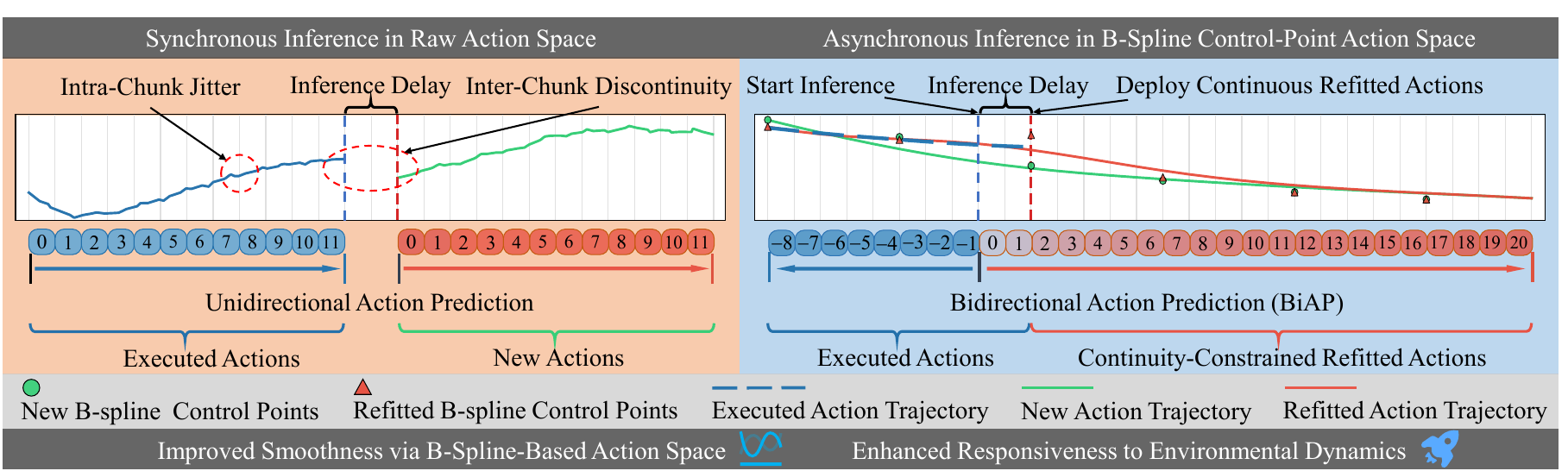}
\captionof{figure}{Overview of ABPolicy.
Our method enables real-time and smooth control via: (1) asynchronous inference to prevent execution stalls; (2) a flow-based trajectory generator predicting B-spline control points for intra-chunk smoothness; and (3) bidirectional prediction and refitting to ensure inter-chunk continuity.
}
\label{IdeaPic2}
\end{strip}

\begin{abstract}

Robotic manipulation requires policies that are smooth and responsive to evolving observations. 
However, synchronous inference in the raw action space introduces several challenges, including intra-chunk jitter, inter-chunk discontinuities, and stop-and-go execution.
These issues undermine a policy's smoothness and its responsiveness to environmental changes.
We propose ABPolicy, an asynchronous flow‑matching policy that operates in a B‑spline control‑point action space.
First, the B‑spline representation ensures intra‑chunk smoothness.
Second, we introduce bidirectional action prediction coupled with refitting optimization to enforce inter‑chunk continuity. 
Finally, by leveraging asynchronous inference, ABPolicy delivers real-time, continuous updates.
We evaluate ABPolicy across seven tasks encompassing both static settings and dynamic settings with moving objects. 
Empirical results indicate that ABPolicy reduces trajectory jerk, leading to smoother motion and improved performance.
Project website: \url{https://teee000.github.io/ABPolicy/}.

\end{abstract}

\section{INTRODUCTION}

Robotic manipulation in real-world environments demands control policies that are both temporally smooth and responsive to evolving observations~\cite{brohan2022rt, zhao2023learning, liu2024rdt}. 
In recent years, there has been growing interest in applying action chunking techniques~\cite{zhao2023learning} and diffusion-based models~\cite{chi2023diffusion,black2410pi0} to robot manipulation within imitation learning frameworks.
While this paradigm has achieved impressive performance, existing methods, which operate in the raw action space with synchronous inference, face several challenges:
(1) intra-chunk jitter that degrades the smoothness of action trajectories; 
(2) inter‑chunk discontinuities that introduce jerks at chunk boundaries and induce distribution shift in subsequent observations; 
(3) stop‑and‑go execution caused by synchronous inference, which undermines responsiveness to dynamic environmental changes.

These considerations bring us to the core question: how can we find an action space or representation that guarantees trajectory smoothness while being easy to integrate into asynchronous inference? 
Recent efforts have explored multiple avenues to mitigate this challenge~\cite{studer2024factorized,habekost2024inverse}.
ACT~\cite{zhao2023learning} and SmolVLA~\cite{shukor2025smolvla} use temporal ensembling to suppress jitter by weighted-averaging multiple rollouts, but the averaged action is not guaranteed to be valid~\cite{black2025real}.
HumanMAC~\cite{chen2023humanmac} and FAST~\cite{pertsch2025fast} parameterize actions using discrete cosine transform (DCT)~\cite{ahmed2006discrete} coefficients, which improves intra‑chunk smoothness but does not resolve discontinuities at chunk boundaries.
PTP~\cite{torne2025learning} predicts past‑action tokens and samples multiple trajectories from the same observation at test time, then selects the one most consistent with prior actions.
This improves temporal coherence but demands repeated sampling during inference, hurting real‑time responsiveness.
BEAST~\cite{zhou2025beast} discretizes B‑spline~\cite{gordon1974b} control points to represent actions, improving smoothness but sacrificing fitting accuracy due to discretization.
In contrast, we use continuous B‑spline control points to achieve lower fitting error.
We then employ a flow‑matching model~\cite{lipman2022flow,liu2022flow} to predict these control points, yielding more accurate action predictions.

Real-world environments are dynamic, requiring agents to act continuously while perceiving evolving observations~\cite{kim2025fine,wen2025tinyvla,yang2025efficientvla,zhao2023learning}.
Conventional synchronous inference induces action stalls that degrade responsiveness~\cite{chi2023diffusion,liu2024rdt,xia2025cage}.
By deploying asynchronous model inference on cloud GPUs, SmolVLA~\cite{shukor2025smolvla} reduces the computational burden on the local device, but at the cost of introducing additional latency from data transmission.
Asynchronous methods like RTC~\cite{black2025real} employ an ``inpainting" strategy to enforce continuity, which utilizes gradient-based guidance and is further enhanced by a weighted fusion of action chunks. 
Yet, this combined mechanism perturbs the learned dynamics and introduces sensitive hyperparameters.
Our approach is orthogonal, we ensure the continuity of asynchronously predicted trajectories via a simple optimization of the B-spline control points.

To achieve both real-time and smooth robotic manipulation, we propose ABPolicy, an asynchronous B-spline flow policy. 
ABPolicy operates in a B-spline control-point action space, generating continuous trajectories for action chunks. 
This representation inherently guarantees intra-chunk smoothness and eliminates jitter. 
To enforce continuity between chunks, we introduce two key mechanisms: bidirectional action prediction, which jointly models a short window of past and future actions, and a continuity-constrained refitting optimization.
This optimization locally adjusts the initial control points of a new trajectory to align with the executed actions. 
Finally, ABPolicy performs inference asynchronously; the manipulator continues to act while the policy updates in the background, enabling real-time adjustments that enhance responsiveness to environmental dynamics.
To validate our approach, we demonstrate its effectiveness on a suite of benchmarks, including three challenging dynamic manipulation tasks and four static ones.

Our main contributions are summarized as follows:
\begin{itemize}
    \item We propose ABPolicy, a flow matching policy that generates action trajectories in a B-spline control point space for inherent smoothness.
    \item We introduce a simple yet effective continuity optimization mechanism, combining bidirectional prediction and continuity-constrained refitting, to seamlessly stitch together asynchronously generated trajectories.
    \item Evaluations on seven manipulation tasks, including three dynamic tasks, confirm that ABPolicy delivers smoother and more reactive control compared to prior methods.
\end{itemize}

\begin{figure*}[t]
\centering
\includegraphics[width=\textwidth, trim=7 247 277 5, clip]{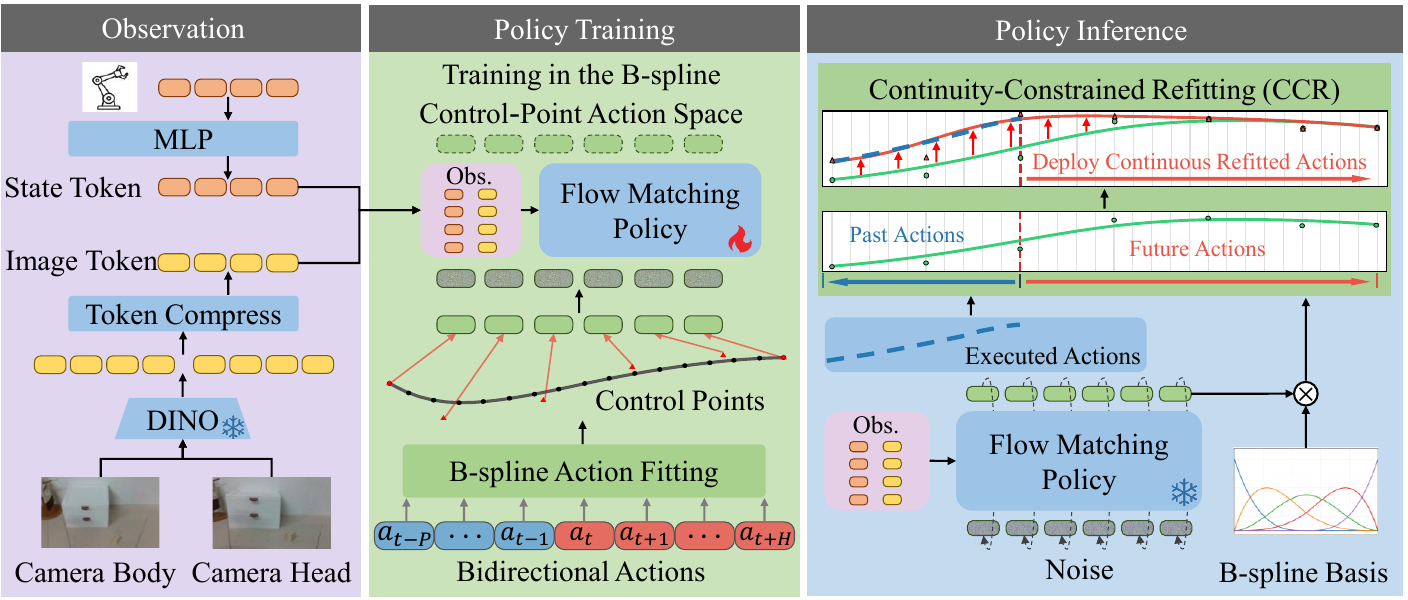}
\caption{Overview of ABPolicy. 
A policy trained for Bidirectional Action Prediction (BiAP) asynchronously generates future B-spline control points at inference time.
These are then optimized by our Continuity-Constrained Refitting (CCR) module to guarantee smooth continuity with the executed trajectory.
}
\label{framework}
\end{figure*}

\section{RELATED WORKS}

\subsection{Action Representation}

Action representation serves as the critical bridge between learned models and robotic motion~\cite{li2023generalist,belkhale2024rt,liu2025hybridvla,von2024art}. 
Researchers have proposed a spectrum of action parameterization strategies. 
One prominent class employs high-level semantic representations.
For example, specifying sub-tasks in natural language~\cite{ahn2022can} or encoding behavior via discrete action keypoints~\cite{di2024keypoint}. 
This family of methods integrates naturally with large-scale foundation models~\cite{wang2024qwen2,beyer2024paligemma}.
However, these approaches typically rely on dedicated low-level motion controllers for execution, which limits their generality and adaptability across tasks and platforms.

Another paradigm directly predicts low-level control signals from observations and human instructions~\cite{brohan2022rt,zitkovich2023rt,lee2024behavior}. 
A common instantiation converts continuous actions into discrete tokens, enabling generation with autoregressive sequence models~\cite{mu2023embodiedgpt,wang2024emu3}. 
For example, straightforward per-dimension binning can be used to discretize actions~\cite{kim2024openvla}. 
However, such discretization introduces quantization error and information loss, which is detrimental to fine-grained manipulation and precise control.

A complementary line of research eschews discretization by predicting continuous low-level control signals from observations and instructions~\cite{team2024octo,zhao2023learning,wen2025dexvla,ke20243d,Ze2024DP3,yan2025maniflow}. 
These methods typically employ U-Net~\cite{ronneberger2015u} or transformer~\cite{vaswani2017attention,peebles2023scalable} architectures to model rich, multimodal action distributions within diffusion- or flow-based generative frameworks. 
By preserving the fidelity of high-frequency motor commands and avoiding quantization artifacts, they deliver superior performance in fine-grained manipulation.
Orthogonal to these approaches, our method represents raw action trajectories with a B-spline parameterization and trains a flow-matching model to generate the continuous B-spline control points.

\subsection{Real-Time Execution}

Within the imitation learning framework, most methods use synchronous inference~\cite{chi2023diffusion,xia2025cage}: the robot waits for the model to finish inference before executing the next action chunk. 
This design introduces latency that weakens responsiveness in dynamic environments~\cite{chi2024universal}. 
As model size increases, computational cost rises, making the delay more pronounced. 
To mitigate this, prior work follows two main paths: 
(1) accelerating diffusion-based or flow-based policies with DDIM sampling~\cite{song2020denoising}, model distillation, or training one-step or few-step denoisers~\cite{geng2025mean,frans2024one} to accelerate the denoising process;
(2) adopting asynchronous inference, which runs policy computation in parallel with control signal dispatch to avoid execution pauses~\cite{shukor2025smolvla,gao2025towards,black2025real}. 
We adopt an asynchronous inference scheme, running model inference and robot control dispatch in separate threads. 
This design eliminates idle time and maintains real-time responsiveness to environmental changes.

\section{METHOD}

Our design targets two goals: smooth action trajectories and asynchronous real-time inference.
We learn a continuous distribution over B-spline control points using a flow matching model. 
During inference, we asynchronously update a subset of these points to ensure the forthcoming trajectory remains continuous with recently executed actions.

\subsection{B-Spline Trajectory Parameterization}

We employ B-splines~\cite{gordon1974b} to parameterize action trajectories, which ensures smoothness and provides a compact representation. 
Specifically, we use cubic B-splines (degree $p=3$). This choice guarantees $C^2$ continuity, meaning both the velocity (first derivative) and acceleration (second derivative) of the action trajectory are continuous. 
Such smoothness is crucial for generating physically realistic motions and avoiding the abrupt changes that can arise from predicting raw actions.
In practice, this process is applied independently to each dimension of the action vector.

Given a discrete action sequence $\{a_t\}_{t=0}^{T-1}$, where $t$ denotes the time step, the objective is to determine a set of $N$ control points $\{c_i\}_{i=0}^{N-1}$ that optimally approximates the sequence. A B-spline curve $s(t)$ of degree $p$ is defined as follows:
\begin{equation}
    s(t) = \sum_{i=0}^{N-1} c_i N_{i,p}(t),
    \label{eq:bspline_def}
\end{equation}
where $N_{i,p}(t)$ is the $i$-th B-spline basis function of degree $p$, defined over a knot vector. The optimal control points $\{c_i^*\}$ are found by minimizing the sum of squared errors between the original actions $a_t$ and the spline curve $s(t)$ evaluated at the corresponding discrete time steps:
\begin{equation}
    \{c_i^*\}_{i=0}^{N-1} = \underset{\{c_i\}}{\arg\min} \sum_{t=0}^{T-1} \left( a_t - s(t) \right)^2.
    \label{eq:bspline_fit}
\end{equation}
This formulation results in a linear least-squares problem, which can be solved efficiently to yield the optimal control points $\{c_i^*\}$.

Once the optimal control points $\{c_i^*\}$ have been obtained from the fitting process, the continuous and smooth action trajectory $\hat{a}(t)$ can be fully reconstructed using the B-spline definition.

\subsection{Bidirectional Action Prediction via Flow Matching}

To generate the action trajectories, we introduce a policy network formulated as a conditional model $\pi_\theta(C_t^* | o_t)$. 
Here, $\theta$ represents the network's trainable parameters, $o_t$ is the observation at time $t$, and $C_t^*=[c_0^*,c_1^*,...,c_{N-1}^* ]$ is the vector of ground-truth control points that parameterize the action chunk. 
This approach fundamentally shifts the learning objective: instead of predicting raw actions, our policy learns to generate a compact set of control points, inherently ensuring the smoothness of the resulting action trajectory.

To enhance the continuity between past and future actions and explicitly models the temporal structure of actions, we adopt a bidirectional action prediction scheme (BiAP). 
At each time step $t$, the policy's target is not a single action $a_t$, but a full action chunk $A_t$ that spans $P$ previous steps and $H$ future steps:
\begin{equation}
    A_t = [a_{t-P}, \dots, a_{t-1}, a_t, a_{t+1}, \dots, a_{t+H-1}].
    \label{eq:action_chunk}
\end{equation}

Our approach bypasses the direct regression of raw action chunks $A_t$ by instead working with their B-spline representation.  
The optimal control points $C_t^*$ corresponding to an action chunk $A_t$ are determined through the least-squares fitting procedure (Eq.~\eqref{eq:bspline_fit}).
These control points subsequently form the regression target for our policy network.

We formulate the prediction of B-spline control points as a conditional generative modeling problem, which we address using flow matching~\cite{lipman2022flow,liu2022flow}. 
Rather than regressing the control points directly, the policy network is trained to predict a conditional vector field that governs a dynamic transformation. 
This transformation maps samples from a simple prior (e.g., a standard Gaussian) to the target distribution of valid control points, conditioned on the observation.
A key advantage of this generative formulation is its inherent ability to model multi-modal trajectory distributions~\cite{chi2023diffusion,black2410pi0}.
The network parameters $\theta$ are optimized via the loss function:
\begin{equation}
    \mathcal{L}_{\text{FM}}^{\tau}(\theta) = \mathbb{E}_{\tau, o_t, z, C_t^*} \left[ \left\| \pi_\theta(C_t^\tau | o_t) - (C_t^* - z) \right\|^2 \right],
    \label{eq:flow_matching_loss_revised}
\end{equation}
where $C_t^*$ denotes the ground-truth control points for observation $o_t$, $z \sim \mathcal{N}(0, \mathbf{I})$ is sampled from the prior, $\tau \sim U[0, 1]$, and $C_t^\tau = (1-\tau)z + \tau C_t^*$ is a point on the linear path between the noise and data.
Minimizing this objective compels the network's output to align with the target vector $(C_t^* - z)$, thus learning the vector field that transforms noise into the desired data, conditioned on $o_t$.

\begin{figure}[t]
\centering
\includegraphics[width=\columnwidth, trim=5 346 400 5, clip]{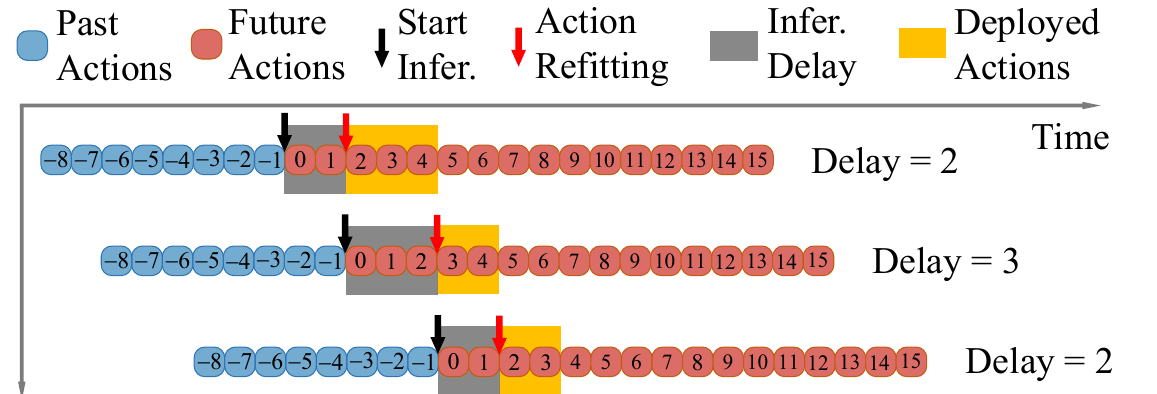}
\caption{
Asynchronous inference overview. 
During inference delay, the robot executes the prior cycle’s actions.
Gray shaded regions denote the inference-delay window, whereas orange shaded regions indicate the actions being executed.
}
\label{Async}
\end{figure}

\subsection{Continuity-Constrained Refitting}

Asynchronous inference design enhances the system's real-time responsiveness to dynamic changes~\cite{black2025real,shukor2025smolvla,gao2025towards}. 
However, this architecture introduces an unavoidable latency between when an observation is captured and when the corresponding new action trajectory is available. 
As shown in Fig.~\ref{IdeaPic2} and Fig.~\ref{Async}, during the model inference period, the robot continues to execute actions from the previously computed trajectory. 
As a result, a direct application of the newly predicted trajectory would cause action discontinuity, as it fails to account for the actions executed during the delay.

To address this challenge, we introduce the Continuity-Constrained Refitting (CCR) mechanism, illustrated in Fig.~\ref{framework}.
The core idea is to locally adjust the initial segment of the newly predicted trajectory to enforce continuity with the sequence of executed actions.

Let $\{c_{\text{pred},i}\}_{i=0}^{N-1}$ be the set of control points predicted by the policy network. Let $\{a_t^{\text{exec}}\}_{t=0}^{P-1}$ be the sequence of $P$ actions that were executed. 
Our goal is to derive a new set of control points, $\{c_{\text{new},i}\}_{i=0}^{N-1}$, that forms a smooth continuation from this history.

To this end, we exploit the local support property of B-splines. The refitting procedure is constrained to adjust only the initial $N_{\text{free}}$ control points, while the subsequent points from the policy's prediction, $\{c_{\text{pred},i}\}_{i=N_{\text{free}}}^{N-1}$, remain unaltered. 
The optimal values for this initial ``free" subset are determined by solving the following least-squares problem:
\begin{equation}
    \{c_{\text{new},i}\}_{i=0}^{N_{\text{free}}-1} = \underset{\{c_i\}_{i=0}^{N_{\text{free}}-1}}{\arg\min} \sum_{t=0}^{P-1} \left( a_t^{\text{exec}} - \hat{s}_{\text{new}}(u_t) \right)^2,
    \label{eq:ccr_objective}
\end{equation}
where $u_t$ maps the step $t$ to the B-spline domain, and $\hat{s}_{\text{new}}(u)$ is the partially updated trajectory defined as:
\begin{equation}
    \hat{s}_{\text{new}}(u) = \underbrace{\sum_{i=0}^{N_{\text{free}}-1} c_i N_{i,p}(u)}_{\text{Free points (to be optimized)}} + \underbrace{\sum_{i=N_{\text{free}}}^{N-1} c_{\text{pred},i} N_{i,p}(u)}_{\text{Fixed points (from policy)}}.
    \label{eq:ccr_spline_def}
\end{equation}
This optimization minimizes the error between the start of the new trajectory and the executed action history. By solving for the initial control points that best fit this history, CCR effectively ``anchors" the new trajectory to the immediate past, guaranteeing continuity and smoothness.

\subsection{Asynchronous Inference}

To achieve real-time reactivity in dynamic environments, we introduce an asynchronous inference framework that decouples model inference from action execution. As depicted in Fig.~\ref{Async}, this architecture runs inference and control in two parallel threads. While the model computes the next action sequence, the robot executes the trajectory from the prior cycle, effectively hiding inference latency.

Once a new sequence is inferred, we instantly update the action queue by applying our refitting method. This guarantees a smooth transition from the previous action plan, prevents jitter, and immediately triggers the next inference cycle. This asynchronous loop enables the robot to remain responsive to environmental changes without being stalled by computation delays.

\begin{figure*}[t]
\centering
\includegraphics[width=\textwidth, trim=5 110 25 5, clip]{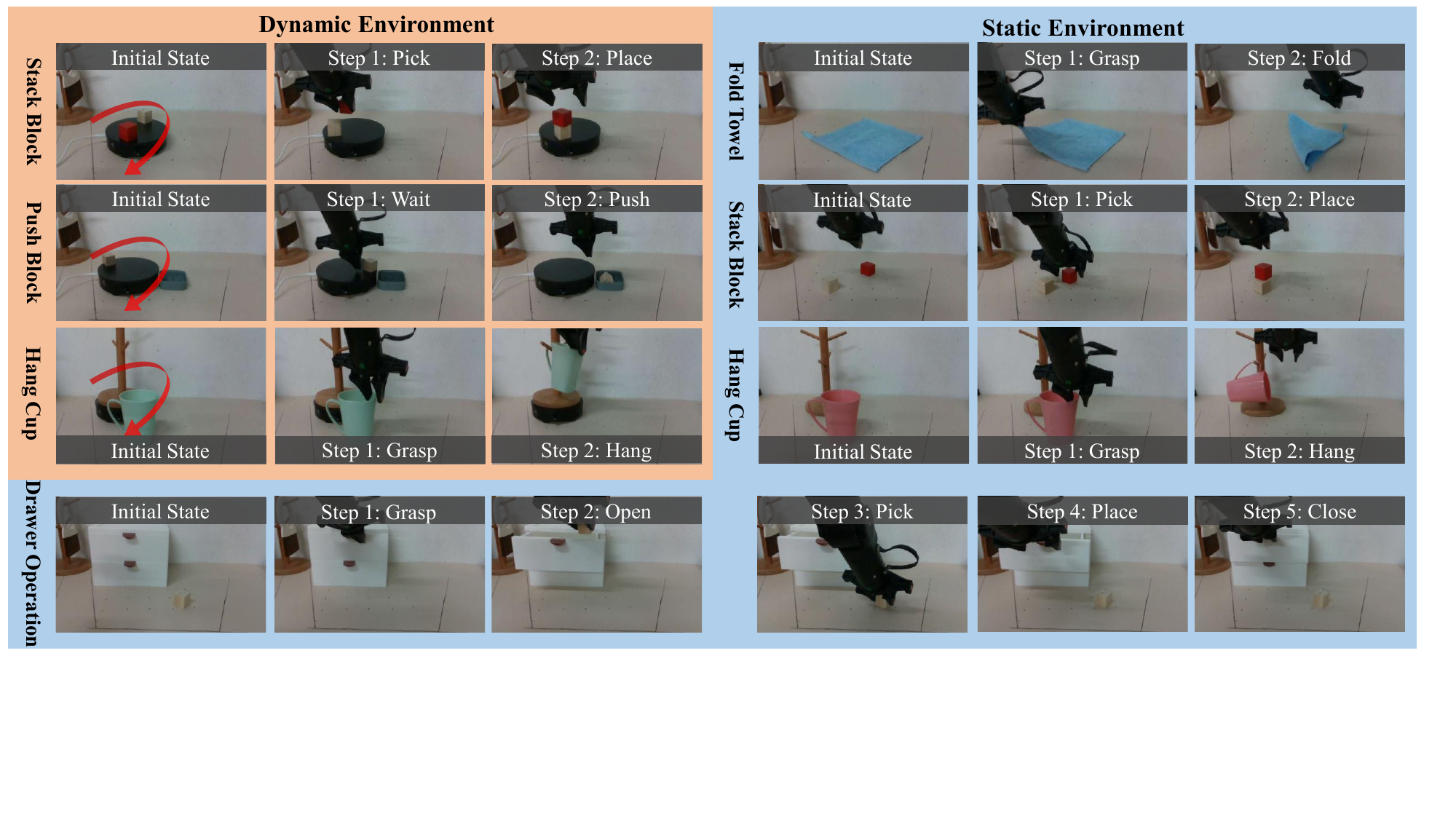}
\caption{
Manipulation tasks in static and dynamic settings, where dynamic tasks involve an object on a platform rotating at a constant velocity (approximately 10 seconds per revolution).
}
\label{TaskPic}
\end{figure*}

\begin{table*}[h]
\centering
\caption{Comparison of synchronous and asynchronous inference performance across dynamic and static tasks.}
\label{tab:main_result}

\begin{tabular}{@{}l ccc @{\hspace{2em}} cccc cccc@{}}
\toprule

& \multicolumn{3}{c}{{Dynamic Tasks}} & \multicolumn{8}{c}{{Static Tasks}} \\
\cmidrule(r){2-4} \cmidrule(l){5-12}

& Stack Block & Push Block & Hang Cup & \multicolumn{2}{c}{Stack Block} & \multicolumn{2}{c}{Fold Towel} & \multicolumn{2}{c}{Hang Cup} & \multicolumn{2}{c}{Drawer Operation} \\
\cmidrule(r){2-2} \cmidrule(r){3-3} \cmidrule(r){4-4} \cmidrule(r){5-6} \cmidrule(r){7-8} \cmidrule(r){9-10} \cmidrule(l){11-12}
 & Success $\uparrow$ & Success $\uparrow$ & Success $\uparrow$ & Success $\uparrow$ & Time $\downarrow$ & Success $\uparrow$ & Time $\downarrow$ & Success $\uparrow$ & Time $\downarrow$ & Success $\uparrow$ & Time $\downarrow$ \\
\midrule
DP~\cite{chi2023diffusion} & 40 & 75 & 35 & 25 & 10.2 & 25 & 12.2 & 85 & 10.8 & \textbf{100} & 18.9  \\
 Sync  & 30          & 75 & 40 & \textbf{85} & 8.7 & 55 & 10.2 & \textbf{90} & 10.1 & \textbf{100} & 17.5 \\
 Async & \textbf{55} & \textbf{85} & \textbf{60} & 80 & \textbf{7.5} & \textbf{60} & \textbf{8.4} & 85 & \textbf{8.2} & \textbf{100} & \textbf{15.8} \\
\bottomrule
\end{tabular}
\end{table*}

\section{EXPERIMENTS}

\subsection{Tasks}

As illustrated in Fig.~\ref{TaskPic}, we evaluate seven manipulation tasks comprising dynamic and static settings.
The dynamic tasks are: stacking a block on a rotating platform, pushing a block on a rotating platform, and hanging a cup onto a rotating rack.
The rotary fixture maintains a constant angular velocity of one revolution every 10 s.
The static tasks are: folding a towel, stacking a block on a stationary base, hanging a cup onto a fixed hook, and putting a block into a drawer. 
For dynamic tasks, we collect 100 demonstrations per task. 
For static tasks, we collect 200 demonstrations per task with the manipulated object randomly initialized within a $20 cm \times 20 cm$ workspace region.
Success rates for each task are calculated based on 20 evaluation runs.

\subsection{Implementation Details}

We employ a 6-DoF AgileX Piper manipulator equipped with a parallel gripper and two fixed third-person Intel RealSense D435 RGB cameras. 
The action space comprises arm joint angles and a continuous gripper aperture command, executed at a 30 Hz control rate.
We fit an independent cubic B-spline (degree $p=3$) for each action dimension, employing an open-uniform (clamped) knot vector.
For BiAP, we set $H=32$ (future) and $P=8$ (history).
Inference is performed on an NVIDIA RTX 4070 Ti Super GPU with 10 denoising steps for the flow-matching model.
The policy network employs a DiT architecture~\cite{peebles2023scalable}, which fuses observations via cross-attention~\cite{liu2024rdt}. 
The observation encoder has two streams: a frozen, pre-trained DINO-V2 model~\cite{oquab2023dinov2} processes the current image frame, while an MLP encodes the robot state from the past 8 frames.
More information can be found on our project website.

\subsection{Comparison of Synchronous and Asynchronous Inference}

To assess how inference mode affects robotic manipulation, we compared synchronous (sync) and asynchronous (async) approaches across seven tasks, categorized as dynamic or static  (Tab.~\ref{tab:main_result}).

For dynamic tasks, the advantage of async inference is more pronounced due to its ability to respond to environmental changes. 
Success rates improved by an average of 18.3\% across the three dynamic tasks.
Under synchronous inference, the robot idles during inference, limiting reactivity to changes and increasing failure rates.
In contrast, asynchronous inference yields lower action latency, enabling timely and effective interaction with moving objects.

For static tasks, the principal benefit of async inference is improved efficiency.
It reduces completion time by 14.2\% on average, with an inference delay of roughly 90 ms.
This efficiency gain is beneficial for real-world deployment, where throughput is a key metric.

\subsection{Analysis of Action Representation Accuracy}

The fidelity of action representation is a critical factor that profoundly influences the performance of robotic manipulation tasks; inaccurate or coarse representations can introduce fitting errors that manifest as suboptimal execution. 
To quantify this effect, we compare four prevalent techniques by evaluating reconstruction accuracy on 40-timestep action chunks, averaged over 200 random samples.
\begin{itemize}
    \item \textbf{Discrete Bins (256):} Quantizes the action space into 256 bins~\cite{kim2024openvla,brohan2022rt,zitkovich2023rt,zhen20243d}.
    \item \textbf{DCT Coefficients (8):} Uses the 8 lowest-frequency DCT coefficients to encode the trajectory~\cite{chen2023humanmac}.
    \item \textbf{B-spline (Discrete):} Represents each action chunk with 8 B-spline control points quantized into 256 bins~\cite{zhou2025beast}.
    \item \textbf{B-spline (Continuous):} Represents each action chunk with 8 continuous B-spline control points.
\end{itemize}

We assessed performance using two metrics.
Mean Error is the average discrepancy between the original and reconstructed trajectories. 
SNR measures fidelity by comparing the power of the original signal to that of the fitting error.

\begin{table}[h!]
\centering
\caption{Comparison of reconstruction accuracy for different action representation methods.}
\label{tab:action_representation}
\begin{tabular}{@{}lcc@{}}
\toprule
{Action Representation Method} & {Mean Error $\downarrow$} & {SNR (dB) $\uparrow$} \\ 
\midrule
256 Discrete Bins~\cite{kim2024openvla}              & 0.0020              & 41.9              \\
DCT Low-Frequency Coeffs~\cite{chen2023humanmac}     & 0.0010              & 44.6              \\
B-spline (Discrete)~\cite{zhou2025beast}            & 0.0025              & 37.7              \\
\textbf{B-spline (Continuous)} (Ours) & \textbf{0.00031}    & \textbf{50.7}     \\ 
\bottomrule
\end{tabular}
\end{table}

As shown in Tab.~\ref{tab:action_representation}, the continuous B‑spline representation attains a mean error of 0.00031 and an SNR of 50.7 dB, outperforming other methods.
This indicates the most faithful, low-noise reconstruction of the original action trajectory, enabling smoother and more effective manipulation.

Notably, the DCT-based approach provides a reasonable approximation but remains less accurate than the continuous B-spline. Standard 256-bin quantization performs worse, exposing the limits of direct discretization. 
Quantizing B-spline control points does not outperform alternatives in reconstruction error.
This underscores the need to maintain continuity in the control-point space; discretization there markedly degrades precision.

\begin{figure}[t]
\centering
\includegraphics[width=\columnwidth, trim=5 5 5 5, clip]{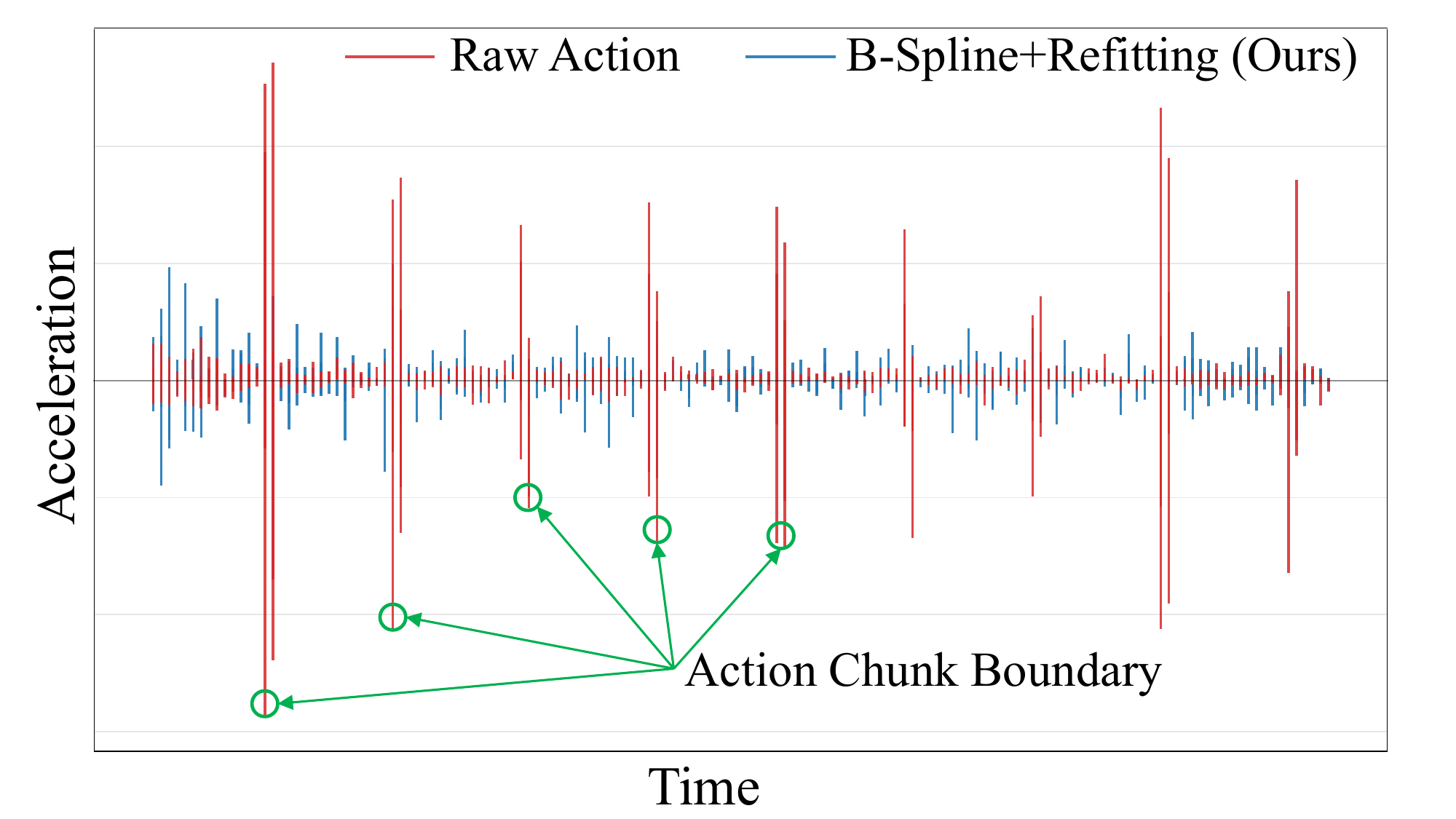}
\caption{
Acceleration comparison between raw actions and our proposed method.
The raw action representation produces large accelerations at the boundaries of action chunks, which leads to jitter.
In contrast, our B-spline representation with refitting reduces boundary spikes and improves smoothness.
}
\label{AccCompareRAW}
\end{figure}

\subsection{Analysis of Action Smoothness}

To evaluate the performance of the B‑spline representation in reducing jitter, we conducted comparative experiments against raw actions.
We used two key metrics:
\begin{itemize}
    \item \textbf{Zero-Crossing Rate (ZCR) of velocity:}  the frequency with which a joint’s velocity changes sign. Lower ZCR indicates fewer oscillations and smoother motion.
    \item \textbf{95th percentile of acceleration (Acc p95):} captures high-frequency changes in motion; lower values indicate less aggressive acceleration and smoother control.
\end{itemize}

\begin{table}[h]
\centering
\caption{Comparison of action representations with trajectory jitter.}
\label{tab:jitter_compare}
\begin{tabular}{@{}lcccc@{}}
\toprule
\multicolumn{1}{c}{{Joint}} & \multicolumn{2}{c}{{Average ZCR of Velocity $\downarrow$}} & \multicolumn{2}{c}{{Acc p95 $\downarrow$}} \\ \cmidrule(lr){2-3} \cmidrule(lr){4-5}
\multicolumn{1}{c}{} & {B-spline} & {Raw} & {B-spline} & {Raw} \\
\midrule
j0 & 0.1535 & 0.2748 & 2.34 & 8.73 \\
j1 & 0.1063 & 0.1615 & 3.63 & 8.04 \\
j2 & 0.1321 & 0.1334 & 3.06 & 7.44 \\
j3 & 0.1422 & 0.1356 & 3.09 & 6.29 \\
j4 & 0.1387 & 0.2201 & 2.35 & 4.23 \\
j5 & 0.1075 & 0.1769 & 3.88 & 8.03 \\
\midrule
\textbf{Average} & \textbf{0.1301} & \textbf{0.1837} & \textbf{3.06} & \textbf{7.13} \\ \bottomrule
\end{tabular}
\end{table}

As shown in Tab.~\ref{tab:jitter_compare}, the B-spline method reduces the average velocity ZCR by 29.2\% compared to raw actions. Furthermore, it achieves a 57.1\% reduction in Acc p95 relative to raw actions.
These results confirm that our method produces smoother actions with lower velocity and acceleration variations and is highly effective at reducing jitter.

\begin{figure}[t]
\centering
\includegraphics[width=\columnwidth, trim=5 195 275 5, clip]{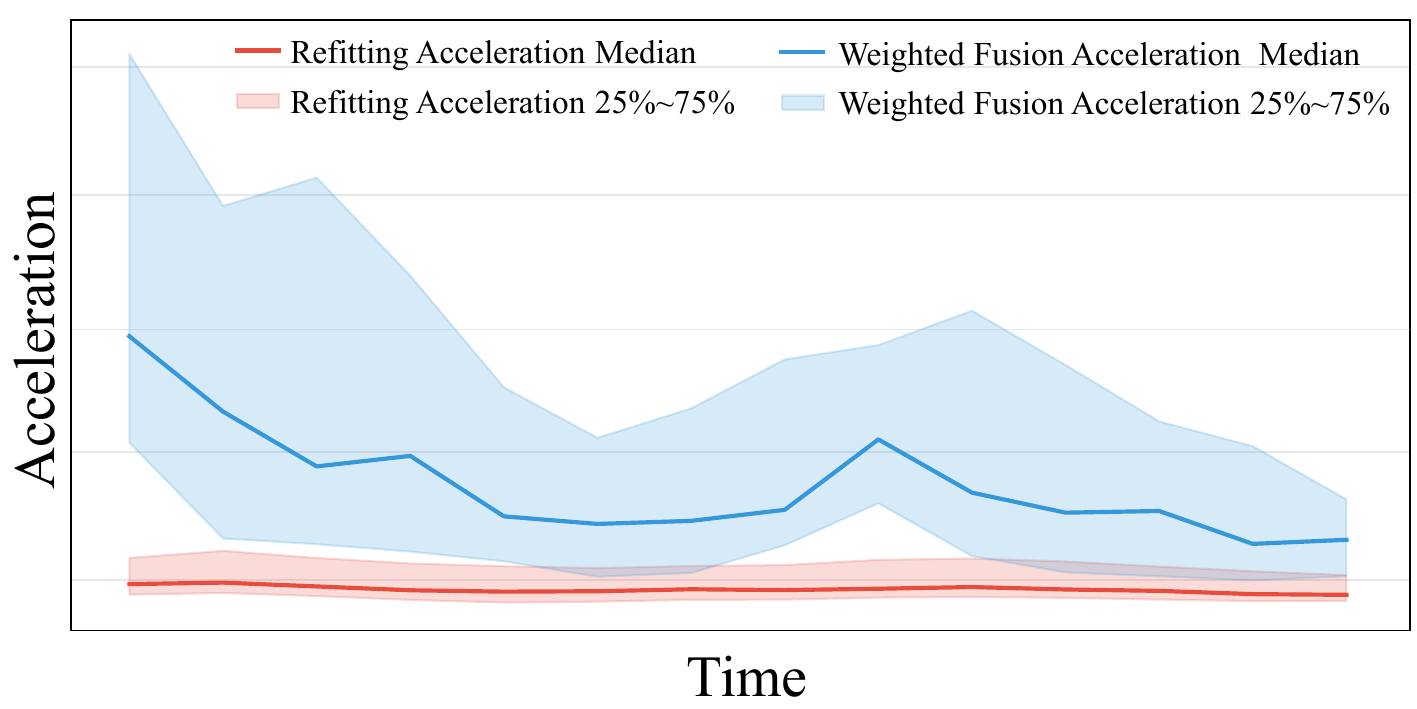}
\caption{
Comparison of action-chunk boundary smoothing methods: our B‑spline refitting versus the weighted fusion. The y‑axis shows acceleration magnitude, and the x‑axis represents the smoothing time window.
}
\label{AccCompareFuse}
\end{figure}

To validate our continuity-constrained refitting method for reducing discontinuities at action chunk boundaries, we benchmarked it against two baselines: raw actions and the weighted fusion technique used in SmolVLA~\cite{shukor2025smolvla} and RTC~\cite{black2025real}.
As illustrated in Fig.~\ref{AccCompareFuse} and Fig.~\ref{AccCompareRAW},  our refitting method yields a markedly smoother action profile. 
We model the action trajectory using a cubic B-spline of degree $p=3$~\cite{gordon1974b}. 
This approach provides a mathematical guarantee of $C^2$ continuity, ensuring the resulting velocity and acceleration profiles are inherently smooth.

\subsection{Analysis of Bidirectional Action Prediction}

\begin{table}[h]
\centering
\caption{
Ablation study on the effectiveness of BiAP, comparing success rate against the discontinuity metric at action boundaries, measured before and after refitting.
}
\label{tab:biap_ablation}
\begin{tabular}{lccc}
\toprule
 & {Success Rate (\%) $\uparrow$} & {Initial Jitter $\downarrow$} & {Refitted Jitter $\downarrow$} \\
\midrule
w/o BiAP        & 60                             & 0.0220                               & 0.0180                                \\
w/ BiAP         & \textbf{85}                    & \textbf{0.0170}                      & \textbf{0.0097}                       \\
\bottomrule
\end{tabular}
\end{table}

Our ablation study on the static block stacking task (Tab.~\ref{tab:biap_ablation}) confirms the efficacy of BiAP. Its inclusion boosts the success rate from 60\% to 85\%, a performance gain stemming from enhanced motion continuity. To quantify this, we measure the transitional jitter between action segments. BiAP reduces the initial jitter by nearly 23\% and enables our refitting module to cut the final jitter by 46\% relative to the baseline (0.0097 vs. 0.018). These findings demonstrate that BiAP is crucial for generating smooth, low-jitter trajectories.

\section{CONCLUSION}

This paper introduced ABPolicy, an asynchronous B-spline flow policy that resolves critical smoothness and latency issues in robotic manipulation. 
By combining a B-spline representation with bidirectional prediction and refitting, our method ensures smooth, continuous trajectories, while its asynchronous design enables the real-time responsiveness crucial for dynamic tasks. 
Experimental results confirm ABPolicy reduces jerk and improves manipulation performance, offering a powerful framework for developing more agile and capable robots for real-world environments.




\bibliographystyle{IEEEtran}
\bibliography{ref}

\end{document}